\documentclass{article}

\usepackage{PRIMEarxiv}
\usepackage{xcolor}
\usepackage[utf8]{inputenc} 
\usepackage[T1]{fontenc}    
\usepackage{hyperref}       
\usepackage{url}            
\usepackage{booktabs}       
\usepackage{amsfonts}       
\usepackage{nicefrac}       
\usepackage{microtype}      
\usepackage{lipsum}
\usepackage{multirow}
\usepackage{fancyhdr}       
\usepackage{graphicx}       
\graphicspath{{media/}}     

\title{The Role of Language Models in Modern Healthcare: A Comprehensive Review}

\author{
  Amna Khalid \\
  CHRISTUS Santa Rosa Hospital \\
  New Braunfels, TX\\
  \And
  Ayma Khalid \\
  Riphah International University \\
  Lahore, Pakistan \\
  \And
  Umar Khalid\\
  Palo Alto, CA
}

\begin{document}
\maketitle

\begin{abstract}
The application of large language models (LLMs) in healthcare has gained significant attention due to their ability to process complex medical data and provide insights for clinical decision-making. These models have demonstrated substantial capabilities in understanding and generating natural language, which is crucial for medical documentation, diagnostics, and patient interaction. This review examines the trajectory of language models from their early stages to the current state-of-the-art LLMs, highlighting their strengths in healthcare applications and discussing challenges such as data privacy, bias, and ethical considerations. The potential of LLMs to enhance healthcare delivery is explored, alongside the necessary steps to ensure their ethical and effective integration into medical practice.
\end{abstract}


\section{Introduction}
Deep learning has revolutionized the way we understand human behavior, emotions, and healthcare-related challenges~\cite{shi2022multiscale, yu2023modality, li2021micro, hong2019characterizing}. In recent years, breakthroughs in clinical language processing have paved the way for transformative changes in the healthcare industry. These advancements hold great promise for the deployment of intelligent systems that can support decision-making, accelerate diagnostic workflows, and enhance the quality of patient care. Such systems have the potential to assist healthcare professionals as they navigate the growing body of medical knowledge, interpret complex patient records, and craft individualized treatment plans. The promise of these systems has generated significant excitement within the healthcare community \cite{he2023survey, wang2023large, yu2023leveraging}.

The power of large language models (LLMs) lies in their ability to analyze vast amounts of medical literature, patient data, and the rapidly growing body of clinical research. Healthcare data \cite{peng2022learning, peng2023generating} is inherently intricate, heterogeneous, and extensive. LLMs function as critical tools that help alleviate information overload for healthcare professionals. By automating the processing of medical texts, extracting key insights, and applying the knowledge, LLMs have the potential to drive significant research breakthroughs and improve patient care, contributing meaningfully to the evolution of the medical field.

The excitement surrounding LLMs is largely driven by the impressive capabilities of advanced models like OpenAI’s GPT-3.5, GPT-4 \cite{brown2020language, openai2023gpt4}, and Google’s Bard. These models have shown remarkable proficiency across a broad range of natural language understanding tasks, underscoring their pivotal role in healthcare applications. With their ability to comprehend and generate human-like text, these models are set to have a transformative impact on healthcare, where accurate communication and information management are paramount \cite{zhang2023one}.

Natural language processing (NLP) has undergone significant advancements, with each milestone building on the strengths and limitations of previous approaches. Early developments, such as recurrent neural networks (RNNs), laid the groundwork for contextual understanding in NLP tasks. However, their limitations in handling long-range dependencies became clear, necessitating new approaches in the field.

The turning point came with the introduction of the Transformer architecture, which effectively addressed the challenge of capturing distant relationships between words. This innovation was crucial for the development of more advanced NLP models. The advent of sophisticated language models such as Llama 2 \cite{touvron2023llama} and GPT-4, both of which benefit from extensive training datasets, has propelled NLP to new heights, allowing for deeper understanding and near-human-level text generation.

Within healthcare, specialized versions of models like BERT, including BioBERT and ClinicalBERT \cite{lee2020biobert, huang2019clinicalbert}, were developed to address the unique challenges of clinical language, such as medical terminology, ambiguity, and variability in usage. However, the use of LLMs in the highly sensitive healthcare sector requires careful consideration of privacy, security, and ethics. Patient data must be rigorously protected, and models must be designed to avoid perpetuating biases or causing harm. Despite these challenges, the potential for LLMs to improve healthcare outcomes and drive innovation remains a key focus of ongoing research and development.

This review serves as a comprehensive guide for medical researchers and healthcare professionals aiming to optimize the use of LLMs in their practices. It provides a detailed exploration of LLM technologies, their applications in healthcare, and critical discussions on fairness, bias, privacy, transparency, and ethical considerations. By addressing these aspects, this review highlights the importance of integrating LLMs into healthcare in a responsible, equitable, and effective manner to maximize benefits for both patients and providers.

The paper is organized into the following sections:
\begin{itemize}
    \item \textbf{Section 2} introduces the fundamental architecture of LLMs, including key components such as Transformers, foundational models, and their multi-modal capabilities.
    \item \textbf{Section 3} explores the application of LLMs in healthcare, detailing their various use cases and the performance metrics used to evaluate them in clinical environments.
    \item \textbf{Section 4} delves into the challenges that LLMs face in healthcare, focusing on issues such as explainability, security, bias, and ethical concerns.
    \item Finally, the paper concludes with a summary of the findings, discussing the transformative potential of LLMs while addressing the need for careful implementation to mitigate limitations and ethical challenges.
\end{itemize}
\section{Overview of Large Language Models}
\label{sec:overview}

Large language models (LLMs) have rapidly advanced due to their ability to understand and generate human-like text across a variety of natural language processing (NLP) tasks~\cite{petroni2019language, brown2020language}. These models are distinguished by their extensive number of parameters, pre-training on vast text datasets, and subsequent fine-tuning for specific tasks~\cite{radford2018improving, chowdhery2022palm, touvron2023llama}. In this section, we examine the core architecture of LLMs, highlight key examples, and explore pre-training methodologies as well as the role of transfer learning~\cite{radford2019language}.

LLMs leverage the Transformer architecture, which excels in capturing long-range dependencies within text~\cite{vaswani2017attention}. The self-attention mechanism inherent to this architecture enables models to focus on different parts of the input text based on their relevance, improving the handling of complex linguistic relationships.

\subsection{Transformers and Their Role in Language Models}

A hallmark of LLMs is their scale~\cite{fedus2022switch, du2022glam}, pre-training on immense text corpora~\cite{wang2022pre, touvron2023llama}, and the fine-tuning process tailored to particular tasks~\cite{wei2021finetuned}. These models, composed of billions of parameters, are designed to recognize intricate patterns in language data. After undergoing broad pre-training, they are refined using smaller, task-specific datasets, resulting in enhanced performance across a variety of NLP applications.

The introduction of the Transformer framework revolutionized the field by addressing the limitations of earlier architectures like recurrent neural networks (RNNs)~\cite{vaswani2017attention}. This evolution led to the development of powerful models like GPT-4~\cite{openai2023gpt4} and Llama 2~\cite{touvron2023llama}, significantly improving natural language understanding and generation.

\subsection{Multi-Modal Language Models: Expanding Capabilities}
A significant progression in AI is the rise of multi-modal language models (MLLMs), which integrate data from multiple sources, such as text, images, and audio. These models, such as BLIP-2~\cite{li2023blip}, extend the traditional capabilities of LLMs by incorporating multiple modalities, allowing for more versatile and robust outputs~\cite{wu2023visual}. MLLMs enable tasks such as visual question answering (VQA) and cross-modal content generation, opening up new possibilities for real-world applications.

\begin{table*}[!t]
\centering
\caption{Summary of Multi-Modal Language Models}
\label{tab:multimodal_models}
\begin{tabular}{|p{3cm}|c|p{5cm}|p{5cm}|}
\hline
Model & Year & Capabilities & Applications \\ \hline
BLIP-2 \cite{li2023blip} & 2023 & Image-text integration using Qformer & Visual question answering, image-text retrieval \\ \hline
Visual ChatGPT \cite{wu2023visual} & 2023 & Text and image interaction via GPT & Complex queries requiring visual inputs \\ \hline
MoVA \cite{zong2024mova} & 2024 & Mixture of experts for image and text & Multi-modal content generation and analysis \\ \hline
\end{tabular}
\end{table*}

\subsection{Applications of Large Language Models in Healthcare}

LLMs have also become prominent in healthcare, where they support tasks such as medical diagnostics, patient care, and drug discovery~\cite{liu2021ai, datta2022bert}. Tailored models like BioBERT~\cite{lee2020biobert} and ClinicalBERT~\cite{huang2019clinicalbert} are designed to handle the specialized language found in medical records and research. Newer models, including GPT-4 and Google's Bard, are setting new benchmarks in medical question answering and related healthcare applications~\cite{wang2023large}.

\begin{table*}[!t]
\centering
\caption{Overview of Large Language Models in Healthcare}
\label{tab:papers_table}
\begin{tabular}{|p{2.2cm}|c|p{3.8cm}|p{3.5cm}|p{1.6cm}|}
\hline
Model           & Year & Use Case                                                                                    & Institution                                                & Source Code                                          \\ \hline

BioMistral \cite{labrak2024biomistral}       & 2024 & Medical Question Answering                                                              & Avignon Université, Nantes Université                                  &  \href{https://huggingface.co/BioMistral/BioMistral-7B}{model}                                                    \\ \hline

Med-PaLM 2 \cite{singhal2023towards}       & 2023 & Medical Question Answering                                                              & Google Research, DeepMind                                  &                                                      \\ \hline
Radiology-Llama2 \cite{liu2023radiology} & 2023 & Radiology Imaging Analysis                                                              & University of Georgia                                      &                                                      \\ \hline
DeID-GPT \cite{liu2023deid}         & 2023 & Data De-identification                                                                  & University of Georgia                                      & \href{https://github.com/yhydhx/ChatGPT-API}{code}                \\ \hline
Med-HALT \cite{umapathi2023med}    & 2023 & Hallucination Detection                                                                 & Saama AI Research  & \href{https://github.com/medhalt/medhalt}{code}                  \\ \hline                        
ChatCAD \cite{zhao2023chatcad+}         & 2023 & Computer-Aided Diagnosis                                                                & ShanghaiTech University                                    & \href{https://github.com/zhaozh10/ChatCAD}{code}                  \\ \hline
BioGPT \cite{luo2022biogpt}        & 2023 & Classification, Relation Extraction, Question Answering                                 & Microsoft Research                                         & \href{https://github.com/microsoft/BioGPT}{code}                  \\ \hline
GatorTron \cite{yang2022gatortron}        & 2022 & Medical Textual Similarity, Inference, Question Answering                               & University of Florida                                               & \href{https://github.com/uf-hobi-informatics-lab/GatorTron}{code} \\ \hline
\end{tabular}
\end{table*}

\subsection{Real-World Healthcare Applications of Large Language Models}
LLMs have been widely adopted across various healthcare functions, with applications continuing to expand rapidly. These models assist in clinical decision-making, analysis of medical records, and improving patient interactions~\cite{dasgupta2022language}. The vast capability of LLMs to process medical data offers benefits in areas such as diagnostics, administrative efficiency, and overall healthcare delivery~\cite{li2023text, agbavor2022predicting}.

\begin{itemize}
\item \textbf{Medical Diagnostics:} LLMs can help physicians diagnose illnesses by analyzing patient data, including symptoms and medical histories, to identify potential health conditions~\cite{chen2023boosting}.
\item \textbf{Patient Care:} Through personalized recommendations and ongoing patient monitoring, LLMs improve the quality of patient care by providing real-time insights~\cite{ali2023using}.
\item \textbf{Clinical Decision Support:} LLMs offer healthcare professionals evidence-based recommendations, enhancing clinical decision-making and treatment strategies~\cite{matin2023leveraging}.
\item \textbf{Medical Literature Review:} By summarizing large volumes of medical literature, LLMs help healthcare professionals stay current with new developments and best practices~\cite{sallam2023utility}.
\item \textbf{Drug Discovery:} LLMs facilitate drug discovery by analyzing molecular data to identify potential compounds for new drugs~\cite{liu2021ai, uludougan2022exploiting}.
\item \textbf{Virtual Health Assistants:} LLMs serve as the backbone for healthcare chatbots that provide continuous health monitoring and medical advice~\cite{bill2023fine}.
\item \textbf{Radiology and Imaging:} Multi-modal LLMs assist radiologists by analyzing imaging data and improving diagnostic precision~\cite{ma2023cephgpt}.
\item \textbf{Automated Report Generation:} LLMs automate the generation of medical reports from diagnostic images, speeding up workflows in radiology and pathology~\cite{zhao2023chatcad+}.
\end{itemize}

\begin{table*}[!t]
\centering
\caption{Evaluation Metrics for LLMs in Healthcare Applications}
\label{tab:eval_metrics}
\begin{tabular}{|p{2cm}|c|p{4cm}|p{5cm}|}
\hline
\textbf{Metric} & \textbf{Task} & \textbf{Description} & \textbf{Key Results} \\ \hline
Perplexity & Language Generation & Measures model uncertainty & Lower perplexity indicates better language generation performance \\ \hline
BLEU & Translation & Evaluates overlap between generated and reference text & ClinicalGPT achieved a BLEU score of 13.9~\cite{wang2023clinicalgpt} \\ \hline
ROUGE & Summarization & Assesses recall of generated summaries & BioMedLM attained a ROUGE-L score of 24.85~\cite{li2023huatuo} \\ \hline
F1 Score & Classification & Combines precision and recall for a balanced metric & GatorTron obtained an F1 score of 0.9627 for medical relation extraction~\cite{yang2022gatortron} \\ \hline
\end{tabular}
\end{table*}

\subsection{Performance Metrics and Model Comparisons}
Benchmarking LLM performance is crucial for assessing their effectiveness across different healthcare tasks. Commonly used benchmarks, such as MMLU (Massive Multitask Language Understanding) and HumanEval, evaluate LLMs on various tasks, including problem-solving and code generation~\cite{hendrycks2020measuring, chen2021evaluating}. Table \ref{tab:llm_perf_bench} presents a comparison of several state-of-the-art models based on these benchmarks.

\begin{table*}[!t]
\centering
\caption{Benchmark Comparison of Large Language Models}
\label{tab:llm_perf_bench}
\begin{tabular}{|p{2cm}|c|p{4cm}|p{3cm}|c|}
\hline
Model & MMLU Score & HumanEval (Coding) & Release Date \\ \hline
GPT-4 Turbo & 86.4 & 85.4 & April 2024 \\ \hline
Claude 3.5 & 88.7 & 92.0 & June 2024 \\ \hline
Llama 3 & 86.1 & 81.7 & March 2024 \\ \hline
Gemini Ultra & 83.7 & 74.3 & December 2023 \\ \hline
\end{tabular}
\end{table*}

\section{Challenges and Future Directions}
\label{sec:challenges}

The incorporation of large language models (LLMs) in healthcare is not without obstacles. These hurdles include the need for greater transparency in model decisions, ensuring data privacy and security for sensitive patient information, addressing biases to guarantee fairness, preventing the generation of false or misleading outputs, and establishing regulatory frameworks for ethical AI use in medical contexts. Overcoming these challenges is vital for fully harnessing LLMs' potential to improve healthcare while maintaining ethical and legal standards.

\subsection{Improving Model Transparency and Interpretability}

One significant challenge when applying LLMs in healthcare is their lack of interpretability. These models often function as "black boxes," making it difficult for healthcare providers to understand how specific recommendations or predictions are generated. This lack of clarity can hinder adoption, as medical professionals require transparent decision-making processes to ensure accuracy and trust. In healthcare, where every decision must be well-founded, the opaque nature of LLMs is particularly problematic. To address this, efforts are underway to develop more interpretable models that offer insight into their decision-making processes, fostering trust in AI-generated recommendations~\cite{ali2023chatgpt, reddy2023evaluating}. Enhancing transparency and interpretability remains a key research focus in healthcare AI~\cite{briganti2023clinician, bisercic2023interpretable, jiang2023balanced}.

\subsection{Data Privacy and Security Risks}

When applied in healthcare settings, LLMs handle vast amounts of sensitive information, including personally identifiable data. Ensuring this data is processed and stored securely, in compliance with privacy regulations, is a significant challenge. One concern is the unintentional exposure of personal health information (PHI) during the training process, which could lead to privacy violations. Furthermore, the ability of LLMs to infer sensitive information from anonymized data presents additional privacy risks~\cite{omiye2023large}. To mitigate these threats, it is essential to implement robust anonymization techniques, secure data storage, and compliance with ethical guidelines, ensuring that patient data remains protected throughout the use of LLMs in healthcare~\cite{thapa2023chatgpt}.

\subsection{Ensuring Fairness and Reducing Bias}

LLMs can inherit biases from the data they are trained on, particularly if the datasets include unequal representations of demographic groups or healthcare outcomes. These biases can lead to disparities in medical recommendations and outcomes, which can be harmful in clinical settings. Researchers must develop strategies to identify, reduce, and prevent biases within these models, ensuring that LLMs contribute to equitable healthcare solutions. Ongoing audits and evaluations are critical for identifying and mitigating biases in both training data and model outputs~\cite{thapa2023chatgpt}. Collaboration between domain experts, data scientists, and ethicists can foster the development of fair and unbiased AI in healthcare.

\subsection{Preventing Hallucinations in Medical AI}

LLMs sometimes generate false or misleading information—commonly referred to as hallucinations—which can be particularly dangerous in healthcare applications where accuracy is critical. These models may produce plausible-sounding, but factually incorrect, content without providing traceable sources~\cite{tian2023opportunities}. Healthcare professionals must be cautious when using LLMs, validating AI-generated content to avoid the risks associated with incorrect medical guidance. Current research is focused on addressing these hallucination challenges, with benchmarks like Med-HALT being developed to evaluate how well models perform in medical reasoning and information retrieval~\cite{umapathi2023med}.

\subsection{Legal, Ethical, and Regulatory Frameworks}

The use of LLMs in healthcare also raises significant legal and ethical questions. Issues such as the generation of sensitive or distressing medical content, or the potential for spreading misinformation, necessitate strict regulatory oversight. Furthermore, there are concerns about plagiarism, impersonation, and the overall integrity of LLM-generated content. Regulatory frameworks, such as the EU’s AI Act and the U.S. HIPAA, provide essential guidelines for the safe and responsible deployment of AI in healthcare~\cite{omiye2023large, novelli2024generative}. These laws ensure patient data protection and set ethical standards for the use of AI technologies in sensitive environments, fostering trust and accountability in AI-powered healthcare.

\section{Closing Remarks}

The adoption of large language models in healthcare presents substantial opportunities for enhancing medical decision-making and information retrieval. These models, equipped with advanced capabilities, have the potential to improve workflows and patient outcomes across various healthcare applications. However, realizing their full potential requires overcoming key challenges such as ensuring model transparency, protecting sensitive data, reducing biases, and preventing erroneous outputs. As researchers and practitioners continue to collaborate, the focus must remain on developing ethical, trustworthy, and fair AI systems that meet the rigorous standards of healthcare. Continued innovation, combined with careful consideration of ethical and regulatory concerns, will shape the future of LLMs in medical practice.

\begin{table*}[!t]
\centering
\caption{Overview of Challenges and Mitigation Strategies for LLMs in Healthcare}
\label{tab:challenges_table}
\begin{tabular}{|p{3cm}|p{4cm}|p{6cm}|}
\hline
\textbf{Challenge} & \textbf{Impact} & \textbf{Proposed Solution} \\ \hline
Transparency & Lack of understanding in AI-generated decisions & Develop interpretable models and provide decision explanations \\ \hline
Data Security & Risk of exposing sensitive patient information & Use advanced anonymization and secure data storage protocols \\ \hline
Bias & Perpetuation of unfair treatment outcomes & Conduct regular bias audits and collaborate with domain experts \\ \hline
Hallucinations & Creation of inaccurate or misleading content & Implement rigorous validation and specialized benchmarks like Med-HALT \\ \hline
Ethical and Legal Concerns & Risk of misuse and data breaches & Comply with regulations such as HIPAA and the AI Act, and ensure ethical use of AI \\ \hline
\end{tabular}
\end{table*}

 \newpage

\bibliographystyle{unsrt}
\bibliography{references}

\begin{thebibliography}{10}

\bibitem{shi2022multiscale}
Henglin Shi, Wei Peng, Haoyu Chen, Xin Liu, and Guoying Zhao.
\newblock Multiscale 3d-shift graph convolution network for emotion recognition from human actions.
\newblock {\em IEEE Intelligent Systems}, 37(4):103--110, 2022.

\bibitem{yu2023modality}
Hao Yu, Xu~Cheng, Wei Peng, Weihao Liu, and Guoying Zhao.
\newblock Modality unifying network for visible-infrared person re-identification.
\newblock In {\em Proceedings of the IEEE/CVF International Conference on Computer Vision}, pages 11185--11195, 2023.

\bibitem{li2021micro}
Yante Li, Wei Peng, and Guoying Zhao.
\newblock Micro-expression action unit detection with dual-view attentive similarity-preserving knowledge distillation.
\newblock In {\em 2021 16th IEEE International Conference on Automatic Face and Gesture Recognition (FG 2021)}, pages 01--08. IEEE, 2021.

\bibitem{hong2019characterizing}
Xiaopeng Hong, Wei Peng, Mehrtash Harandi, Ziheng Zhou, Matti Pietik{\"a}inen, and Guoying Zhao.
\newblock Characterizing subtle facial movements via riemannian manifold.
\newblock {\em ACM Transactions on Multimedia Computing, Communications, and Applications (TOMM)}, 15(3s):1--24, 2019.

\bibitem{he2023survey}
Kai He, Rui Mao, Qika Lin, Yucheng Ruan, Xiang Lan, Mengling Feng, and Erik Cambria.
\newblock A survey of large language models for healthcare: from data, technology, and applications to accountability and ethics.
\newblock {\em arXiv preprint arXiv:2310.05694}, 2023.

\bibitem{wang2023large}
Yuqing Wang, Yun Zhao, and Linda Petzold.
\newblock Are large language models ready for healthcare? a comparative study on clinical language understanding.
\newblock {\em arXiv preprint arXiv:2304.05368}, 2023.

\bibitem{yu2023leveraging}
Ping Yu, Hua Xu, Xia Hu, and Chao Deng.
\newblock Leveraging generative ai and large language models: a comprehensive roadmap for healthcare integration.
\newblock In {\em Healthcare}, volume~11, page 2776. MDPI, 2023.

\bibitem{peng2022learning}
Wei Peng, Li~Feng, Guoying Zhao, and Fang Liu.
\newblock Learning optimal k-space acquisition and reconstruction using physics-informed neural networks.
\newblock In {\em Proceedings of the IEEE/CVF Conference on Computer Vision and Pattern Recognition}, pages 20794--20803, 2022.

\bibitem{peng2023generating}
Wei Peng, Ehsan Adeli, Tomas Bosschieter, Sang~Hyun Park, Qingyu Zhao, and Kilian~M Pohl.
\newblock Generating realistic brain mris via a conditional diffusion probabilistic model.
\newblock In {\em International Conference on Medical Image Computing and Computer-Assisted Intervention}, pages 14--24. Springer, 2023.

\bibitem{brown2020language}
Tom Brown, Benjamin Mann, Nick Ryder, Melanie Subbiah, Jared~D Kaplan, Prafulla Dhariwal, Arvind Neelakantan, Pranav Shyam, Girish Sastry, Amanda Askell, et~al.
\newblock Language models are few-shot learners.
\newblock {\em Advances in neural information processing systems}, 33:1877--1901, 2020.

\bibitem{openai2023gpt4}
OpenAI.
\newblock Gpt-4 technical report, 2023.

\bibitem{zhang2023one}
Chaoning Zhang, Chenshuang Zhang, Chenghao Li, Yu~Qiao, Sheng Zheng, Sumit~Kumar Dam, Mengchun Zhang, Jung~Uk Kim, Seong~Tae Kim, Jinwoo Choi, et~al.
\newblock One small step for generative ai, one giant leap for agi: A complete survey on chatgpt in aigc era.
\newblock {\em arXiv preprint arXiv:2304.06488}, 2023.

\bibitem{touvron2023llama}
Hugo Touvron, Thibaut Lavril, Gautier Izacard, Xavier Martinet, Marie-Anne Lachaux, Timoth{\'e}e Lacroix, Baptiste Rozi{\`e}re, Naman Goyal, Eric Hambro, Faisal Azhar, et~al.
\newblock Llama: Open and efficient foundation language models.
\newblock {\em arXiv preprint arXiv:2302.13971}, 2023.

\bibitem{lee2020biobert}
Jinhyuk Lee, Wonjin Yoon, Sungdong Kim, Donghyeon Kim, Sunkyu Kim, Chan~Ho So, and Jaewoo Kang.
\newblock Biobert: a pre-trained biomedical language representation model for biomedical text mining.
\newblock {\em Bioinformatics}, 36(4):1234--1240, 2020.

\bibitem{huang2019clinicalbert}
Kexin Huang, Jaan Altosaar, and Rajesh Ranganath.
\newblock Clinicalbert: Modeling clinical notes and predicting hospital readmission.
\newblock {\em arXiv preprint arXiv:1904.05342}, 2019.

\bibitem{petroni2019language}
Fabio Petroni, Tim Rockt{\"a}schel, Patrick Lewis, Anton Bakhtin, Yuxiang Wu, Alexander~H Miller, and Sebastian Riedel.
\newblock Language models as knowledge bases?
\newblock {\em arXiv preprint arXiv:1909.01066}, 2019.

\bibitem{radford2018improving}
Alec Radford, Karthik Narasimhan, Tim Salimans, Ilya Sutskever, et~al.
\newblock Improving language understanding by generative pre-training.
\newblock 2018.

\bibitem{chowdhery2022palm}
Aakanksha Chowdhery, Sharan Narang, Jacob Devlin, Maarten Bosma, Gaurav Mishra, Adam Roberts, Paul Barham, Hyung~Won Chung, Charles Sutton, Sebastian Gehrmann, et~al.
\newblock Palm: Scaling language modeling with pathways.
\newblock {\em arXiv preprint arXiv:2204.02311}, 2022.

\bibitem{radford2019language}
Alec Radford, Jeffrey Wu, Rewon Child, David Luan, Dario Amodei, Ilya Sutskever, et~al.
\newblock Language models are unsupervised multitask learners.
\newblock {\em OpenAI blog}, 1(8):9, 2019.

\bibitem{vaswani2017attention}
Ashish Vaswani, Noam Shazeer, Niki Parmar, Jakob Uszkoreit, Llion Jones, Aidan~N Gomez, {\L}ukasz Kaiser, and Illia Polosukhin.
\newblock Attention is all you need.
\newblock {\em Advances in neural information processing systems}, 30, 2017.

\bibitem{fedus2022switch}
William Fedus, Barret Zoph, and Noam Shazeer.
\newblock Switch transformers: Scaling to trillion parameter models with simple and efficient sparsity.
\newblock {\em The Journal of Machine Learning Research}, 23(1):5232--5270, 2022.

\bibitem{du2022glam}
Nan Du, Yanping Huang, Andrew~M Dai, Simon Tong, Dmitry Lepikhin, Yuanzhong Xu, Maxim Krikun, Yanqi Zhou, Adams~Wei Yu, Orhan Firat, et~al.
\newblock Glam: Efficient scaling of language models with mixture-of-experts.
\newblock In {\em International Conference on Machine Learning}, pages 5547--5569. PMLR, 2022.

\bibitem{wang2022pre}
Haifeng Wang, Jiwei Li, Hua Wu, Eduard Hovy, and Yu~Sun.
\newblock Pre-trained language models and their applications.
\newblock {\em Engineering}, 2022.

\bibitem{wei2021finetuned}
Jason Wei, Maarten Bosma, Vincent~Y Zhao, Kelvin Guu, Adams~Wei Yu, Brian Lester, Nan Du, Andrew~M Dai, and Quoc~V Le.
\newblock Finetuned language models are zero-shot learners.
\newblock {\em arXiv preprint arXiv:2109.01652}, 2021.

\bibitem{li2023blip}
Junnan Li, Dongxu Li, Silvio Savarese, and Steven Hoi.
\newblock Blip-2: Bootstrapping language-image pre-training with frozen image encoders and large language models.
\newblock In {\em International conference on machine learning}, pages 19730--19742. PMLR, 2023.

\bibitem{wu2023visual}
Chenfei Wu, Shengming Yin, Weizhen Qi, Xiaodong Wang, Zecheng Tang, and Nan Duan.
\newblock Visual chatgpt: Talking, drawing and editing with visual foundation models.
\newblock {\em arXiv preprint arXiv:2303.04671}, 2023.

\bibitem{zong2024mova}
Zhuofan Zong, Bingqi Ma, Dazhong Shen, Guanglu Song, Hao Shao, Dongzhi Jiang, Hongsheng Li, and Yu~Liu.
\newblock Mova: Adapting mixture of vision experts to multimodal context.
\newblock {\em arXiv preprint arXiv:2404.13046}, 2024.

\bibitem{liu2021ai}
Zhichao Liu, Ruth~A Roberts, Madhu Lal-Nag, Xi~Chen, Ruili Huang, and Weida Tong.
\newblock Ai-based language models powering drug discovery and development.
\newblock {\em Drug Discovery Today}, 26(11):2593--2607, 2021.

\bibitem{datta2022bert}
Tanmoy~Tapos Datta, Pintu~Chandra Shill, and Zabir Al~Nazi.
\newblock Bert-d2: Drug-drug interaction extraction using bert.
\newblock In {\em 2022 International Conference for Advancement in Technology (ICONAT)}, pages 1--6. IEEE, 2022.

\bibitem{labrak2024biomistral}
Yanis Labrak, Adrien Bazoge, Emmanuel Morin, Pierre-Antoine Gourraud, Mickael Rouvier, and Richard Dufour.
\newblock Biomistral: A collection of open-source pretrained large language models for medical domains.
\newblock {\em arXiv preprint arXiv:2402.10373}, 2024.

\bibitem{singhal2023towards}
Karan Singhal, Tao Tu, Juraj Gottweis, Rory Sayres, Ellery Wulczyn, Le~Hou, Kevin Clark, Stephen Pfohl, Heather Cole-Lewis, Darlene Neal, et~al.
\newblock Towards expert-level medical question answering with large language models.
\newblock {\em arXiv preprint arXiv:2305.09617}, 2023.

\bibitem{liu2023radiology}
Zhengliang Liu, Yiwei Li, Peng Shu, Aoxiao Zhong, Longtao Yang, Chao Ju, Zihao Wu, Chong Ma, Jie Luo, Cheng Chen, et~al.
\newblock Radiology-llama2: Best-in-class large language model for radiology.
\newblock {\em arXiv preprint arXiv:2309.06419}, 2023.

\bibitem{liu2023deid}
Zhengliang Liu, Xiaowei Yu, Lu~Zhang, Zihao Wu, Chao Cao, Haixing Dai, Lin Zhao, Wei Liu, Dinggang Shen, Quanzheng Li, et~al.
\newblock Deid-gpt: Zero-shot medical text de-identification by gpt-4.
\newblock {\em arXiv preprint arXiv:2303.11032}, 2023.

\bibitem{umapathi2023med}
Logesh~Kumar Umapathi, Ankit Pal, and Malaikannan Sankarasubbu.
\newblock Med-halt: Medical domain hallucination test for large language models.
\newblock {\em arXiv preprint arXiv:2307.15343}, 2023.

\bibitem{zhao2023chatcad+}
Zihao Zhao, Sheng Wang, Jinchen Gu, Yitao Zhu, Lanzhuju Mei, Zixu Zhuang, Zhiming Cui, Qian Wang, and Dinggang Shen.
\newblock Chatcad+: Towards a universal and reliable interactive cad using llms.
\newblock {\em arXiv preprint arXiv:2305.15964}, 2023.

\bibitem{luo2022biogpt}
Renqian Luo, Liai Sun, Yingce Xia, Tao Qin, Sheng Zhang, Hoifung Poon, and Tie-Yan Liu.
\newblock Biogpt: generative pre-trained transformer for biomedical text generation and mining.
\newblock {\em Briefings in Bioinformatics}, 23(6):bbac409, 2022.

\bibitem{yang2022gatortron}
Xi~Yang, Aokun Chen, Nima PourNejatian, Hoo~Chang Shin, Kaleb~E Smith, Christopher Parisien, Colin Compas, Cheryl Martin, Mona~G Flores, Ying Zhang, et~al.
\newblock Gatortron: A large clinical language model to unlock patient information from unstructured electronic health records.
\newblock {\em arXiv preprint arXiv:2203.03540}, 2022.

\bibitem{dasgupta2022language}
Ishita Dasgupta, Andrew~K Lampinen, Stephanie~CY Chan, Antonia Creswell, Dharshan Kumaran, James~L McClelland, and Felix Hill.
\newblock Language models show human-like content effects on reasoning.
\newblock {\em arXiv preprint arXiv:2207.07051}, 2022.

\bibitem{li2023text}
Hongyang Li, Richard~C Gerkin, Alyssa Bakke, Raquel Norel, Guillermo Cecchi, Christophe Laudamiel, Masha~Y Niv, Kathrin Ohla, John~E Hayes, Valentina Parma, et~al.
\newblock Text-based predictions of covid-19 diagnosis from self-reported chemosensory descriptions.
\newblock {\em Communications Medicine}, 3(1):104, 2023.

\bibitem{agbavor2022predicting}
Felix Agbavor and Hualou Liang.
\newblock Predicting dementia from spontaneous speech using large language models.
\newblock {\em PLOS Digital Health}, 1(12):e0000168, 2022.

\bibitem{chen2023boosting}
Zekai Chen, Mariann~Micsinai Balan, and Kevin Brown.
\newblock Boosting transformers and language models for clinical prediction in immunotherapy.
\newblock {\em arXiv preprint arXiv:2302.12692}, 2023.

\bibitem{ali2023using}
Stephen~R Ali, Thomas~D Dobbs, Hayley~A Hutchings, and Iain~S Whitaker.
\newblock Using chatgpt to write patient clinic letters.
\newblock {\em The Lancet Digital Health}, 5(4):e179--e181, 2023.

\bibitem{matin2023leveraging}
Rubeta~N Matin, Eleni Linos, and Neil Rajan.
\newblock Leveraging large language models in dermatology, 2023.

\bibitem{sallam2023utility}
Malik Sallam.
\newblock The utility of chatgpt as an example of large language models in healthcare education, research and practice: Systematic review on the future perspectives and potential limitations.
\newblock {\em medRxiv}, pages 2023--02, 2023.

\bibitem{uludougan2022exploiting}
G{\"o}k{\c{c}}e Uludo{\u{g}}an, Elif Ozkirimli, Kutlu~O Ulgen, Nilg{\"u}n Karal{\i}, and Arzucan {\"O}zg{\"u}r.
\newblock Exploiting pretrained biochemical language models for targeted drug design.
\newblock {\em Bioinformatics}, 38(Supplement\_2):ii155--ii161, 2022.

\bibitem{bill2023fine}
Desir{\'e}e Bill and Theodor Eriksson.
\newblock Fine-tuning a llm using reinforcement learning from human feedback for a therapy chatbot application, 2023.

\bibitem{ma2023cephgpt}
Lei Ma, Jincong Han, Zhaoxin Wang, and Dian Zhang.
\newblock Cephgpt-4: An interactive multimodal cephalometric measurement and diagnostic system with visual large language model.
\newblock {\em arXiv preprint arXiv:2307.07518}, 2023.

\bibitem{wang2023clinicalgpt}
Guangyu Wang, Guoxing Yang, Zongxin Du, Longjun Fan, and Xiaohu Li.
\newblock Clinicalgpt: Large language models finetuned with diverse medical data and comprehensive evaluation.
\newblock {\em arXiv preprint arXiv:2306.09968}, 2023.

\bibitem{li2023huatuo}
Jianquan Li, Xidong Wang, Xiangbo Wu, Zhiyi Zhang, Xiaolong Xu, Jie Fu, Prayag Tiwari, Xiang Wan, and Benyou Wang.
\newblock Huatuo-26m, a large-scale chinese medical qa dataset.
\newblock {\em arXiv preprint arXiv:2305.01526}, 2023.

\bibitem{hendrycks2020measuring}
Dan Hendrycks, Collin Burns, Steven Basart, Andy Zou, Mantas Mazeika, Dawn Song, and Jacob Steinhardt.
\newblock Measuring massive multitask language understanding.
\newblock {\em arXiv preprint arXiv:2009.03300}, 2020.

\bibitem{chen2021evaluating}
Mark Chen, Jerry Tworek, Heewoo Jun, Qiming Yuan, Henrique Ponde De~Oliveira Pinto, Jared Kaplan, Harri Edwards, Yuri Burda, Nicholas Joseph, Greg Brockman, et~al.
\newblock Evaluating large language models trained on code.
\newblock {\em arXiv preprint arXiv:2107.03374}, 2021.

\bibitem{ali2023chatgpt}
Hazrat Ali, Junaid Qadir, Tanvir Alam, Mowafa Househ, and Zubair Shah.
\newblock Chatgpt and large language models (llms) in healthcare: Opportunities and risks.
\newblock 2023.

\bibitem{reddy2023evaluating}
Sandeep Reddy.
\newblock Evaluating large language models for use in healthcare: A framework for translational value assessment.
\newblock {\em Informatics in Medicine Unlocked}, page 101304, 2023.

\bibitem{briganti2023clinician}
Giovanni Briganti.
\newblock A clinician's guide to large language models.
\newblock {\em Future Medicine AI}, (0):FMAI, 2023.

\bibitem{bisercic2023interpretable}
Aleksa Bisercic, Mladen Nikolic, Mihaela van~der Schaar, Boris Delibasic, Pietro Lio, and Andrija Petrovic.
\newblock Interpretable medical diagnostics with structured data extraction by large language models.
\newblock {\em arXiv preprint arXiv:2306.05052}, 2023.

\bibitem{jiang2023balanced}
Yan Jiang, Ruihong Qiu, Yi~Zhang, and Peng-Fei Zhang.
\newblock Balanced and explainable social media analysis for public health with large language models.
\newblock {\em arXiv preprint arXiv:2309.05951}, 2023.

\bibitem{omiye2023large}
Jesutofunmi~A Omiye, Haiwen Gui, Shawheen~J Rezaei, James Zou, and Roxana Daneshjou.
\newblock Large language models in medicine: the potentials and pitfalls.
\newblock {\em arXiv preprint arXiv:2309.00087}, 2023.

\bibitem{thapa2023chatgpt}
Surendrabikram Thapa and Surabhi Adhikari.
\newblock Chatgpt, bard, and large language models for biomedical research: Opportunities and pitfalls.
\newblock {\em Annals of Biomedical Engineering}, pages 1--5, 2023.

\bibitem{tian2023opportunities}
Shubo Tian, Qiao Jin, Lana Yeganova, Po-Ting Lai, Qingqing Zhu, Xiuying Chen, Yifan Yang, Qingyu Chen, Won Kim, Donald~C Comeau, et~al.
\newblock Opportunities and challenges for chatgpt and large language models in biomedicine and health.
\newblock {\em arXiv preprint arXiv:2306.10070}, 2023.

\bibitem{novelli2024generative}
Claudio Novelli, Federico Casolari, Philipp Hacker, Giorgio Spedicato, and Luciano Floridi.
\newblock Generative ai in eu law: liability, privacy, intellectual property, and cybersecurity.
\newblock {\em arXiv preprint arXiv:2401.07348}, 2024.

\end{thebibliography}

\end{document}